\documentclass{article}

    \usepackage[preprint]{neurips_2019}

\usepackage[utf8]{inputenc} 
\usepackage[T1]{fontenc}    
\usepackage{hyperref}       
\usepackage{url}            
\usepackage{booktabs}       
\usepackage{amsfonts}       
\usepackage{nicefrac}       
\usepackage{microtype}      
\usepackage[rgb,dvipsnames]{xcolor}
\usepackage{graphicx}
\usepackage{caption}
\usepackage{subcaption}
\usepackage{bm}
\usepackage{mathrsfs}
\usepackage{multirow}
\usepackage{enumitem}
\title{UniDual: A Unified Model for Image and Video Understanding}

\author{Yufei Wang  \hspace{0.2cm} Du Tran \hspace{0.2cm} Lorenzo Torresani \\
Facebook AI \\
{\tt\small \{yufei22, trandu, torresani\}@fb.com}}

\begin{document}

\maketitle

\begin{abstract}
Although a video is effectively a sequence of images, visual perception systems typically model images and videos separately, thus failing to exploit the correlation and the synergy provided by these two media. While a few prior research efforts have explored the benefits of leveraging still-image datasets for video analysis, or vice-versa, most of these attempts have been limited to pretraining a model on one type of visual modality and then adapting it via finetuning on the other modality. In contrast, in this paper we introduce a framework that enables joint training of a unified model on mixed collections of image and video examples spanning different tasks. 
The key ingredient in our architecture design is a new network block, which we name {\em UniDual}. It consists of a shared 2D spatial convolution followed by two parallel point-wise convolutional layers, one devoted to images and the other one used for videos. For video input, the point-wise filtering implements a temporal convolution. For image input, it performs a pixel-wise nonlinear transformation. 
Repeated stacking of such blocks gives rise to a network where images and videos undergo partially distinct execution pathways, unified by spatial convolutions (capturing commonalities in visual appearance) but separated by point-wise operations (modeling patterns specific to each modality). 
Extensive experiments on Kinetics and ImageNet demonstrate that our UniDual model jointly trained on these datasets yields substantial accuracy gains for both tasks, compared to 1) training separate models, 2) traditional multi-task learning and 3) the conventional framework of pretraining-followed-by-finetuning. On Kinetics, the UniDual architecture applied to a state-of-the-art video backbone model (R(2+1)D-152) yields an additional video@1 accuracy gain of 1.5\%. 
\end{abstract}

\section{Introduction}
\label{sec:intro}
\vspace{-0.4em}


Videos are commonly referred to as ``moving images'' as they differ from still-photos in their ability to capture and render dynamic aspects of the scene. Yet, each video frame is effectively an image. Thus, it is conceivable that computer vision systems may benefit from joint training over both modalities in order to take full advantage of their commonalities and of existing datasets in these two domains. Despite the intuition that image and videos share strong appearance cues and provide synergistic information, most existing machine vision models are designed to operate on only one of these two modalities and thus are typically trained on one form of data only. Indeed, the most prominent architectures in computer vision today are models that were specifically designed either for images (e.g., Inception~\cite{inception}, ResNet~\cite{resnet}, DenseNet~\cite{densenet}) or for videos (e.g., C3D~\cite{c3d}, Two-Stream Networks~\cite{SimonyanZ14}, I3D~\cite{i3d}).

We argue that there are two limitations in modeling images and videos independently. First, video models cannot fully benefit from the many existing large-scale and well-curated image datasets, such as ImageNet~\cite{imagenet}, Places~\cite{places}, or COCO~\cite{coco}, which are the precious result of multiple-year efforts in data collection and annotation by the research community. Similarly, image models cannot take advantage of a growing number of well-annotated video benchmarks~\cite{kinetics, something2, sigurdsson2016hollywood, zhao2019hacs, LiLV18}  due to model discrepancies between the two domains. Second, in many industry settings, image models and video models are separately developed, trained, and deployed even when they address related tasks that may be advantageous to tackle jointly. For example, the task of object classification in images may provide strong cues for action recognition in videos since actions (e.g., playing tennis) can  often be disambiguated on the basis of the objects appearing in the scene (e.g., a tennis racquet). Due to the model differences in the two domains, today such synergies are only suboptimally exploited (e.g., object recognition is typically optimized on images only rather than on images and videos) and often require costly duplication of functionalities and resources. 

Rare exceptions to this separate handling of images and videos are represented by methods that pretrain a model on one modality and then adapt its weights and/or its architecture via fine-tuning on data of the other modality. For example, many early convolutional neural networks (CNNs) for action recognition leveraged CNNs pretrained on still-image datasets to compute frame-based descriptors from the video~\cite{tsn,SimonyanZ14,FeichtenhoferPZ16,FeichtenhoferNIPS16}. More recently, several authors have proposed strategies to take architectures pretrained on images and extend them in a second stage to model temporal information in video, e.g., by adding recurrent units~\cite{lrcn, prernn} by inflating 2D filters to 3D~\cite{i3d}, or by distillation~\cite{distinit}. Various attempts have also been made to transfer knowledge learned from videos to the image domain. Most of this prior work leverages the space-time consistency of moving objects in videos, and performs unsupervised or weakly supervised learning with videos to improve spatial feature representations or object detectors for still-images~\cite{Mobahi2009DeepLF,  Leistner11, Zou2012DeepLO, Alessandro12, Wang_UnsupICCV2015}. We note, however, that this entire genre of methods relies on a two-step training procedure that is aimed at optimizing performance on one modality only by pretraining on the other.

In this paper, we attempt to unify image and video understanding by proposing a {\em single} model that can be {\em jointly} trained on mixed collections of image and video examples in order to leverage the benefits of both types of data and to produce {\em strong performance on both domains}. Reaching this goal hinges on the following question: how can we model the spatial information shared by still and moving images, while also capturing the temporal information contained in video? We address this question by introducing a novel architecture that unifies the modeling of shared spatial appearance information in images and video, but it enables separate (dual) handling of temporal and image-specific information. Hence, we name our model {\em UniDual}.  The key ingredient in our architecture is a new network block that consists of a shared 2D spatial convolution followed by two parallel point-wise convolutional layers, one devoted to images and the other one used for videos. For video input, the point-wise filtering implements a temporal convolution. For image input, it performs a pixel-wise nonlinear transformation. Repeated stacking of such blocks gives rise to a network where images and videos undergo partially distinct execution pathways, unified by spatial convolutions (capturing commonalities in visual appearance) but separated by dual point-wise operations (modeling patterns specific to each modality). Our comparative experiments on Kinetics and ImageNet demonstrate that our proposed scheme yields accuracy gains on both tasks over 1) disjoint training, 2) traditional multi-task learning, and 3) finetuning after pretraining.

In summary, the contributions of this paper are as follows:\vspace{-.1cm}
\begin{itemize}[leftmargin=*,noitemsep,topsep=0pt]
	\item We introduce a unified model that jointly learns from both images and videos. Our {\em UniDual} building block includes a shared spatial 2D convolution learned from both modalities, and two parallel pointwise convolutions learned separately from videos and images.
	\item We investigate the effect of the joint learning on both modalities with an extensive comparative study. We show that the unified model outperforms various baselines, and boosts the performance on both the image task and the video task. 
	\item We also demonstrate that the unified model can be learned from more than two datasets to address several visual tasks simultaneously. 
\end{itemize}

\vspace{-0.4em}
\section{A Unified Model for Image and Video Understanding}
\vspace{-0.5em}
\subsection{Background}
\vspace{-0.4em}
In this subsection we review the Residual Network (ResNet)~\cite{resnet} and the R(2+1)D model~\cite{r2p1d}, which we use as inspiration for our network design.

\begin{figure}[htbp]
\vspace{-0.4em}
\centering
     \begin{subfigure}[t]{0.165\textwidth}
         \includegraphics[width=0.9\columnwidth]{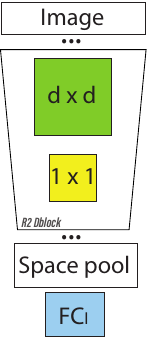}
         \caption{R2D model for images.}
         \label{fig:r2d}
     \end{subfigure}   \hspace{2mm}       
     \begin{subfigure}[t]{0.17\textwidth}
         \includegraphics[width=0.892\columnwidth]{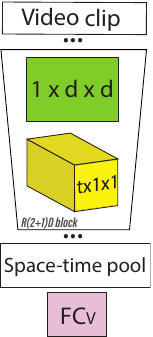}
         \caption{R(2+1)D model for videos.}
         \label{fig:r2p1d}
     \end{subfigure}     \hspace{2mm}   
          \begin{subfigure}[t]{0.29\textwidth}
         \centering
         \includegraphics[width=0.9\columnwidth]{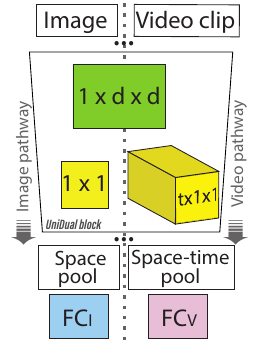}
         \caption{Our {\em UniDual} network for images and videos.}
         \label{fig:unified model}
     \end{subfigure}     \hspace{2mm}   
     \begin{subfigure}[t]{0.29\textwidth}
         \centering
         \includegraphics[width=0.9\columnwidth]{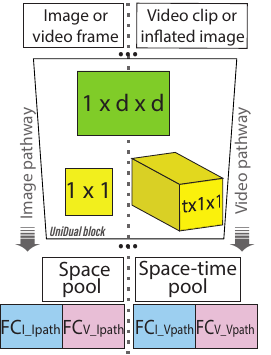}
         \caption{Our {\em UniDual} for images and video with auxiliary losses.}
         \label{fig:unified model bothloss}
     \end{subfigure}    
   \caption{Illustration of different image and video architectures. The blocks are stacked multiple times for all architectures. (a) The R2D model is designed for image task, where the input is a single image, and the residual blocks perform spatial convolutions with 2D or 1D filters. (b) The R(2+1)D model is designed for video tasks. The input is a video clip of $L$ frames, and each block consists of a 3D convolution with 2D spatial filters followed by a 3D convolution with 1D temporal filters. (c) The input to our {\em UniDual} model can be either an image or a video clip. The UniDual block includes a convolutional layer with 2D spatial filters followed by two distinct pointwise convolutional layers, one applied to video input (using $t\times 1 \times 1$ filters) and one applied to images (using $1 \times 1$ filters). (d) We propose to add an auxiliary loss to both the image pathway and the video pathway. This forces all filters to be updated with both modalities, which results in more similar data statistics in all layers and more stable optimization.} 
   \label{fig:architecture}
\vspace{-1em}
\end{figure}

\vspace{-0.8em}
\paragraph {Residual networks.}
ResNet models for image analysis are built by stacking residual blocks:
\begin{equation}
z_i=z_{i-1}+\mathcal{F}(z_{i-1}; {W_i})
\end{equation}
where $z_i$ denotes the activations of the $i$th residual block, and $\mathcal{F}(; W_i)$ is the residual mapping to be learned between block $(i-1)$ and block $i$. The residual mapping typically consists of two convolutional layers with $3\times3$ filters parameterized by $W_i$ with ReLU after each convolution layer. In deep ResNets, the block often has a bottleneck design consisting of three convolutional layers with filters of size $1\times1$, $3\times3$, and $1\times1$, respectively. The purpose of the first and the third $1\times1$ pointwise convolutions is to decrease and then restore the dimensionality in order to reduce the computational cost of the $3\times3$ convolution. In this paper we denote ResNet models for image understanding as R2D. Figure~\ref{fig:r2d} shows an illustration of a modified R2D block, with a $1\times1$ point-wise convolution after the $d\times d$ convolution. We explain this R2D variant in Section~\ref{sec:ablation}

\vspace{-0.8em}
\paragraph {R(2+1)D models.} R(2+1)D are ResNet models designed for video tasks. With the input being a video clip, the activations $z_i$ are 4D tensor $C_i\times L_{i}\times H_{i}\times W_i$, where $C_i$ is the number of channels,  and $L_i$ is the number of frames in the temporal dimension.  The building block of R(2+1)D can be viewed as replacing the traditional 3D convolution used in early video CNNs~\cite{c3d,i3d} with two separate layers, one performing 3D convolution with 2D spatial filters, the second applying a 3D convolution with 1D temporal filters. Specifically, the first layer in the block applies $M_i$ 2D spatial filters of size $C_i\times 1 \times d \times d$, and the second layer employs $C_{i+1}$ 1D point-wise temporal filters of size $M_i \times t \times 1 \times 1$. Typically, $t=3$ is used for temporal dimension, and $d=3$ or $d=1$ is used for the spatial dimension. $M_i$ is the dimension of the intermediate activation after spatial convolution. Note that because both the spatial and the temporal filters are applied via 3D convolutions, the activations remain 4D tensors preserving temporal ordering information. The factorization of temporal and spatial filtering was shown in~\cite{r2p1d} to be advantageous over traditional 3D convolution as  it eases the optimization by forcing the learning of separate spatial and temporal information. The R(2+1)D model is illustrated in Figure~\ref{fig:r2p1d}. 

\vspace{-0.4em}
\subsection{Unified Models for Images and Video}
\vspace{-0.4em}
The R(2+1)D model separates the spatial convolutions and temporal convolutions, which complies with our motivation of learning 1) spatial appearance information jointly for both images and videos and 2) temporal information for videos only. Therefore, we naturally develop our unified model on top of a R(2+1)D backbone. Next, we first introduce a na\"{\i}ve application of R(2+1)D to jointly model images and video and we then present a more sophisticated solution using our UniDual block. 

\vspace{-0.4em}
\paragraph{R(2+1)D jointly trained on videos and inflated images.}
The R(2+1)D model takes video clips of length $L$ as input. Therefore one simple way to jointly model videos and images with a vanilla R(2+1)D model is by inflating images as static videos. This entails taking an input image $I$ and copying it $L$ times to create a static video clip $I_{inflate} = [I, I, ...I]$ that can be fed into the R(2+1)D model. To handle simultaneous joint training with two distinct tasks, separate prediction layers for the image task and the video task are added, as in multi-task learning. Because both videos and images are represented in the form of clips, it is possible to train this model with mini-batches consisting of a mix of both images and videos. The gradient of each input will be computed with respect to the appropriate loss (either the image loss or the video loss) but all layers (except the prediction layers) will be updated with gradients computed for both types of input. 

 
%

\vspace{-0.8em}
\paragraph{The UniDual block.}
The approach outlined above views images as static videos, and trains a shared R(2+1)D model on mixed collections of images and videos. This strategy neglects the essential difference between images and videos: videos include temporal information that is not present in images. The temporal convolutions in the R(2+1)D model is redundant for images.

To give the model the power to learn temporal information for real videos while maintaining shared spatial convolutions between the two modalities, we propose a unified model with building block {\em UniDual}. The block consists of a shared spatial 2D convolution between images and videos, followed by two distinct but parallel point-wise convolutions, one for images and one for videos. The dual pointwise convolutions implement a $t \times 1 \times 1$ temporal convolution for videos and a $1 \times 1$  pixel-wise convolution for images. Under this model, images and videos undergo partially distinct execution pathways, unified by spatial convolutions but separated by point-wise operations that are specific to each modality. The temporal pointwise convolution are updated with gradients computed from video examples, while the $1 \times 1$ filters are updated with gradients computed from image examples only. The 2D convolutional layer is updated with both image and video information. The {\em UniDual} architecture resulting from stacking multiple such blocks is illustrated in Figure~\ref{fig:unified model}. For the final prediction, a pooling layer and a fully connected layer are stacked on top of the residual blocks for both pathways. The pooling for the video pathway is over the spatiotemporal volume, while that for the image pathway is only over the spatial volume. Note that with this model, there is no longer need to artificially inflate images in the form of video clips. 

The $1 \times 1$  convolutions may appear superfluous as the image pathway could in principle be defined by the stacking of the 2D convolutions shared with the video pathway. However, by including $1 \times 1$ convolutions, the image pathway has the same total number of nonlinear activations as the video pathway, rendering the model equally deep for both modalities. Furthermore, the $1 \times 1$  convolution in each block can be viewed as ``aligning'' the output of the image pathway to that of the video pathway before the next shared 2D convolution in the subsequent block. 


%
%

\vspace{-0.8em}
\paragraph{Auxiliary Losses.}
The unified model architecture still suffers from one problem caused by the fact that the dual point-wise convolutional layers are trained on single modalities. Specifically, the $t \times 1 \times 1$ filters are trained only with video data, the $1 \times 1 \times 1$  filters are trained only with image data while the 2D filters in the network are trained with data from both sources. This results in different data distributions for different layers through the network.
Empirically, we found this statistical data differences in the layers to render the optimization very  challenging.

To address this problem, we add an auxiliary loss to each pathway. As shown in Figure~\ref{fig:unified model bothloss}, the layers in the video pathway are trained with clips corresponding to both real video and inflated images by leveraging an auxiliary loss defined for inflated images with respect to the image task (say, classification on ImageNet). Similarly, an auxiliary loss over individual video frames with respect to the video task (say, action recognition on Kinetics) is added to the image pathway. With the addition of these auxiliary losses, all layers in the {\em UniDual} blocks are updated from both sources consistently, while maintaining the ability to learn image-specific features and temporal information for video.

\vspace{-0.8em}
\paragraph{Inference.}
After training, the auxiliary FC layers are removed. At inference time, only one pathway is activated for each input. When the input is an image, the unified model acts as a ResNet 2D model, while for the video input, the unified model turns into a R(2+1)D model. This allows our unified model to operate simultaneously on two different forms of input sources.

\vspace{-0.8em}
\section{Experiments}
\vspace{-0.8em}
We begin by presenting in subsection~\ref{sec:ablation} a comparative study to validate the advantages of the proposed unified model against the R2D and R(2+1)D architectures and several additional variants of these models to allow joint training over images and videos. In order to make this extensive comparison computationally feasible, we limit the depth of all models to 34 layers and we use short clips of 8 frames as input. We refer the reader to subsection~\ref{subsec:soa} for an evaluation using deeper models trained on longer clips, which will demonstrate the state-of-the-art accuracy of our model.

\vspace{-0.5em}
\subsection{Comparative Study}
\vspace{-0.5em}
\label{sec:ablation}

\paragraph{Datasets.}
For the image task, we use the ImageNet~\cite{imagenet} dataset for image classification. There are 1000 categories in this dataset. We use the full set of 1.2M images for training, and the validation set of 50K images for testing. For the video task, we use the Kinetics-400 dataset~\cite{kinetics} for action recognition. It contains 400 human action classes, with 240K training videos, and 20K validation videos. We use the full training set for training, and the validation set for testing. 

\paragraph{Models.}
\vspace{-3mm}
\begin{table}[ht]
\vspace{-0.4em}
\caption{Models considered in our experimental comparison. For each model we list the input type (image and/or video) and the form of training (separate vs joint).}
   \centering
   \bgroup
   \scalebox{0.85}{
\def\arraystretch{1.1}
\begin{tabular}{c|c|c|c}
\hline
\textbf{Training type }          & \textbf{Model}              & \textbf{Image pathway input}    & \textbf{Video pathway input}                  \\ \hline
\multirow{4}{*}{separate} & R2D                & image                  & N/A                                  \\
                          & R(2+1)D            & N/A                    & video clip                           \\
                          & R2D-finetuned      & image                  & N/A                                  \\
                          & R(2+1)D-finetuned  & N/A                    & video clip                           \\ \hline
\multirow{5}{*}{joint}    & R2D-multitask      & image inflated as clip & video clip                           \\
                          & R(2+1)D-multitask  & image inflated as clip & video clip                           \\
                          & R2D-LSTM-multitask & image inflated as clip & video clip                           \\
                          & UniDual            & image                  & video clip                           \\
                          & UniDual-Aux        & image OR video frame   & video clip OR image inflated as clip \\ \hline
\end{tabular}}
\egroup
\label{tbl:ablation_input}
\vspace{-0.3em}
\end{table}

Table~\ref{tbl:ablation_input} lists the models considered in our comparison, which we further discuss here:\vspace{-.1cm}
\begin{itemize}[leftmargin=*,noitemsep,topsep=0pt]
\item \textbf{R2D} and \textbf{R(2+1)D}. These are baseline models separately trained on images and videos, respectively. The R(2+1)D model is trained on Kinetics. The R2D model is built by replacing the $t \times 1 \times 1$ temporal convolution of R(2+1)D with a $1\times1$ convolution, and it is trained on ImageNet.
\item \textbf{R2D-finetuned} and \textbf{R(2+1)D-finetuned}. These are models pretrained on one modality and then finetuned on the other modality. Specifically, R2D-finetuned is obtained by {\em deflating} the R(2+1)D pretrained on Kinetics as a R2D model. This is done by converting each $t \times 1 \times 1$ filter into a $1 \times 1$ kernel by summing up the weights over the $t$ temporal channels. Then, the resulting R2D is finetuned on ImageNet. The model {R(2+1)D-finetuned} is instead obtained by {\em inflating} the R2D pretrained on ImageNet (as in~\cite{i3d}) and by then finetuning it on Kinetics.
\item \textbf{R2D-multitask} and \textbf{R(2+1)D-multitask}. Each of these models is jointly trained on ImageNet and Kinetics as in multi-task learning, with all layers shared except for the distinct final prediction layers. The input is either a single ImageNet image inflated as a clip or an actual Kinetics video clip. The learning objective is a weighted combination of the image classification loss on ImageNet input, and action recognition loss on Kinetics input. We set the loss weights to be equal for the two tasks. Each mini-batch is constructed by randomly sampling examples from both tasks. Since the size of the ImageNet dataset is larger than that of the Kinetics dataset, we use different sampling rates in order to produce mini-batches that contain an equal proportion of examples for both tasks on average. All layers in each of these models are shared across the two tasks, except for the final FC layers. The difference between R2D-multitask and R(2+1)D-multitask is that the former does not include temporal convolutions. Although R2D-multitask takes a clip as input, it separately processes the 8 frames in the clip via 2D convolutions and then performs temporal average pooling over the resulting 8 tensors before the final clip-classification by the FC layer. In~\cite{r2p1d} this model was shown to be superior to a R2D model using 2D filters spanning the entire temporal extent of the clip (as opposed to learning frame-based 2D filters, as we do here). 
\item \textbf{R2D-LSTM}. The architecture of this model is identical to that of R2D-multitask, except that the temporal average pooling is replaced with a single-layer LSTM of 512 hidden units aggregating the information from the 8 tensors produced by the final 2D convolutional layer. This model is also trained in a multitask fashion over the two tasks. Two distinct prediction layers on top of the LSTM perform the classification for the two tasks.
\item \textbf{UniDual} and \textbf{UniDual-Aux}. These two models implement our proposed architecture. The former uses the conventional ImageNet loss and Kinetics loss on the image pathway and video pathway, respectively. The latter uses two auxiliary losses, one for image task and one for video task. 
\end{itemize}

\vspace{-3mm}
\begin{table}[ht]
\caption{Accuracies achieved on the ImageNet and Kinetics benchmarks by different models, all based on the same R(2+1)D backbone of 34 layers (deflated to 2D for R2D). Our unified UniDual-Aux network gives the best performance on both tasks.}
   \centering
   \scalebox{0.85}{
\def\arraystretch{1.1}
\begin{tabular}{c|c|cc|cc}
\hline
\multirow{2}{*}{\textbf{Use of other modality}} & \multirow{2}{*}{\textbf{Model}} & \multicolumn{2}{c|}{\textbf{Kinetics} }                                      & \multicolumn{2}{c}{\textbf{ImageNet}} \\ \cline{3-6} 
                                       &                        & clip@1 (\%) & video@1  (\%)                                    & top 1 (\%)        & top 5 (\%)       \\ \hline \hline
N/A                                    & R2D                    & N/A           & N/A                                                 & 69.2        & 89.2       \\
N/A                                    & R(2+1)D                & 54.7        & 67.3                                              & N/A           & N/A          \\ \hline
\multirow{2}{*}{Pretrain}              & R2D-finetuned          & N/A           & N/A                                                 & 69.6        & 89.2       \\
                                       & R(2+1)D-finetuned      & 55.4        & 67.3                                              & N/A           & N/A          \\ \hline
\multirow{5}{*}{Joint training}        & R2D-multitask          & 56.0        & 67.9                                              & 70.5        & 90.1       \\
                                       & R(2+1)D-multitask      & 57.1        & 69.7                                              & N/A           & N/A          \\
                                       & R2D-LSTM-multitask     & 54.5        & 67.3                                              & 69.2        & 89.1       \\
                                       & UniDual                & 54.6        & 67.0                                              &         70.0      &       89.7      \\
                                       & UniDual-Aux            & \textbf{57.8}     &   \textbf{70.5} & \textbf{70.9}        & \textbf{90.3}       \\ \hline
\end{tabular}
}
\label{tbl:ablation}
\vspace{-0.8em}
\end{table}
\vspace{-0.8em}
\paragraph{Training and Inference.} All experiments in our comparative study are carried out using the same backbone model for fair comparison. We use a R(2+1)D model with 34 layers, taking as input a clip of 8 consecutive RGB frames.  During training, the 8-frame clips are randomly sampled from the original video. The video frames/images are resized to a spatial size of $128 \times 171$ pixels, and then the input to the model is formed by randomly cropping windows of size $112 \times 112$ .We train the models with synchronous distributed SGD on GPU clusters using caffe2 (\cite{caffe2}), with 16 4-GPU machines. We use mini-batches of size 16 per GPU. All the models are learned with a base learning rate of 0.01, and a total of 45 epochs. After the 10 initial warm-up epochs, the learning rate is reduced by a factor of 10 every 10 epochs. Epoch size is set to 1M for single modality, and 2M when training with both modalities. During inference, we resize the video spatially to size $128 \times 171$, and then use the center $112 \times 112$ window. For testing, we sample 10 uniformly spaced-out 8-frame clips from a full-length video, and compute the final prediction by averaging the prediction scores of individual clips.

\vspace{-0.1em}
\paragraph{Results.}
Table~\ref{tbl:ablation} shows the accuracies achieved with different models on Kinetics and ImageNet (median of 3 runs). The first two numerical columns 
report performance on Kinetics. From these numbers we can draw several interesting observations. First, we can notice that R(2+1)D, which is the only variant not using ImageNet as auxiliary training set, yields low accuracy. This underscores the importance of leveraging still-image information for video analysis. We can also observe that {R(2+1)D-multitask} performs better than {R(2+1)D-finetuned}, even though finetuning is more commonly adopted than multitask learning to optimize performance on a target task. A possible explanation is that finetuning causes the the model to ``forget'' the information originally learned from ImageNet, while multi-task learning helps retaining this information and leveraging it more effectively for action recognition. The accuracy gap between {R(2+1)D-multitask} and {R2D-multitask} stresses the importance of properly modeling the temporal information in a video for action recognition. Without auxiliary losses, {UniDual} performs similarly to vanilla R(2+1)D, which does not use ImageNet. However, the addition of auxiliary losses for our model gives a 3.5\% gain in video accuracy compared to {UniDual} and renders {UniDual-Aux} the best model in this comparison. This shows the importance of matching  data statistics for different layers through the network during the optimization. {UniDual-Aux} outperforms {R(2+1)D-multitask} by 0.8\% and {R(2+1)D-finetuned} by 3.2\% in video accuracy. This demonstrates the beneficial effect of the {\em UniDual} block, which enables the learning of shared appearance information from the image and video modalities but decouples the temporal and image-specific information through the dual convolutions.

The last two columns of Table~\ref{tbl:ablation} show top-1 and top-5 accuracy on ImageNet. The relative performance of the different models on this image task matches what seen on the video task. It is interesting to note that the use of Kinetics helps still-image recognition: the accuracy of R2D (without Kinetics) is considerably lower than that achieved by {R2D-finetuned} and especially by {R2D-multitask}. {UniDual-Aux} achieves again the highest accuracy among the models considered here, thus demonstrating the benefits of our unified architecture even for still-image analysis.

We note that, although the gap between UniDual-Aux and the multitask networks is not as large as for the other models, it is consistently present in all our experiments. For example, while Table~\ref{tbl:ablation} shows median accuracy over 3 runs for each model, the min/max clip@1 accuracy over the 3 runs is 57.0/57.1(\%) for R(2+1)D-multitask and 57.8/58.2 (\%) for UniDual-Aux, suggesting a stable gap. Furthermore, we note that R2D-multitask and R(2+1)D-multitask are also variants of our joint-training approach and they have not been studied in prior work.

\subsection{Learning a Unified Model from Several Sources}
\begin{table}[!t]
\vspace{-0.5em}
  \caption{UniDual-Aux enables joint learning of a unified model from several sources spanning several tasks. In this case the sources are: Kinetics, ImageNet and MiniSports. UniDual-Aux yields gains on all three tasks compared to training models separately on each dataset.\vspace{-.1cm}}
   \centering
   \scalebox{0.9}{
\def\arraystretch{1.1}
\begin{tabular}{c|cc|cc|cc}
\hline
\multirow{2}{*}{\textbf{Model}} & \multicolumn{2}{c|}{\textbf{Kinetics}} & \multicolumn{2}{c|}{\textbf{ImageNet}} & \multicolumn{2}{c}{\textbf{MiniSports}} \\ \cline{2-7} 
                       & clip@1 (\%)       & video@1 (\%)     & top 1  (\%)         & top 5   (\%)       & clip@1  (\%)        & video@1  (\%)       \\ \hline
R2D        & N/A           & N/A           & 69.2        & 89.2       & N/A            & N/A            \\
R(2+1)D            & 54.7        & 67.3        & N/A           & N/A          & 31.4         & 60.5         \\
UniDual-Aux            & 59.6        & 71.7        & 69.9        & 90.0       & 32.3         & 61.9        \\ \hline
\end{tabular}
}
\label{tbl:threedatasets}
\vspace{-0.4em}
\end{table}

Our approach can be extended to learn a single model from more than two datasets to address several visual tasks simultaneously. In this section, we show this potential feature using three datasets: ImageNet, Kinetics, and Sports1M. To make the joint training over these three datasets computationally feasible, we randomly subsample Sports1M to the size of Kinetics, but we still use its full validation set for evaluation. We refer to this reduced version of Sports1M as MiniSports.

We follow the training procedure in Section~\ref{sec:ablation}, but increase the epoch size to 3M examples for the joint training. During joint training, the model sees an equal number of examples from the three datasets. We compare UniDual with models trained separately on each dataset. Table~\ref{tbl:threedatasets} shows median results for 3 runs. Compared to the baselines, our unified model yield accuracy gains on all three datasets, with improvements of 0.8\%, 4.4\%, 1.4\% top 1 accuracy on ImageNet, Kinetics, and MiniSports, respectively. We see a relatively smaller gain on ImageNet compared to joint training on ImageNet and Kinetics only. This is due to the fact that we train with two datasets of video modality, and one dataset with image modality. There is opportunity for studying better balancing strategies over multiple sources in future work. At the same time, we can notice that the leveraging MiniSports as an additional source allows UniDual-Aux to further improve over the results for Kinetics in Table~\ref{tbl:ablation}.

\vspace{-0.2em}
\subsection{Evaluation on Kinetics and ImageNet Using UniDual-152}
\label{subsec:soa}
\vspace{-0.2em}

In this subsection we present results obtained by adopting a deeper architecture trained on longer clips in order to demonstrate the gains hold also when we apply our approach to higher-performing models. We also show that our UniDual network achieves state-of-the-art performance on Kinetics.
\vspace{-0.4em}
\paragraph {Training and inference.}
For these experiments, we adopt a R(2+1)D-152 (152 layers) as backbone for our model. We 
 follow the video preprocessing steps described in~\cite{non-local}, but we us4 16-frame clips as input for the model. During training, each clip is formed by taking a random sequence and sampling 16 frames from it, by taking 1 frame every 4 frames. The input frames have a spatial size of $224 \times 224$ pixels, obtained by randomly cropping from a scaled frame whose shorter edge is randomly resized to either 256 or 320 pixels. For image input, the preprocessing is the same, and the image input is then inflated to a static 16-frame clip. Since the input spatial size is  twice that used by the original R(2+1)D model, we modify the model by adding a spatial max pooling layer after the first (2+1)D convolutional block. The max pooling layer uses pooling regions of size $1\times3\times3$, with a spatial stride of 2 and a temporal stride of 1. We use 16 4-GPU machines for training, with mini-batches of size 8 per GPU. 
 We train the model with 45 epochs in total, with a 10 warm-up epochs, followed by a cosine learning rate schedule of 35 epochs. The base learning rate is set to 0.005. Epoch size is the same as in the previous experiments. 
At inference time, for the Kinetics dataset, we follow the procedure described in~\cite{non-local}: for each 16-frame clip, we spatially resize the frames so that the shorter side is rescaled to 256 pixels, and then we perform fully-convolutional inference spatially on 3 clips of spatial size $256 \times 256$  cropped from the rescaled frames. Along the temporal dimension, we sample 10 clips uniformly over the original video, and we average the prediction scores from individual clips to get the final prediction for the video. For the ImageNet, we rescale the shorter edge to 256, and take $224 \times 224$ center crop as input.

\vspace{-3mm}
\begin{table}[ht!]  
  \caption{Accuracies on Kinetics and ImageNet with models based on a R(2+1)D-152 backbone. On Kinetics, UniDual-Aux yields a sensible gain over the strong R(2+1)D-152, achieving accuracy comparable to the state-of-the-art. On ImageNet, the same UniDual-Aux outperforms R2D-152.\vspace{-.2cm}}
\centering
   \scalebox{0.9}{
\def\arraystretch{1.1}
\begin{tabular}{c|cc|cc}
\hline
\multirow{2}{*}{\textbf{Model}  } & \multicolumn{2}{c|}{\textbf{Kinetics} }& \multicolumn{2}{c}{\textbf{ImageNet}} \\ \cline{2-5} 
                         & video@1 (\%)      & top@5 (\%)      & top 1(\%)      & top 5(\%)     \\ \hline
R2D             & N/A           & N/A           & 74.2          & 92.0         \\
R(2+1)D         & 76.0          & 92.2          & N/A           & N/A          \\
UniDual-Aux              & 77.5          & 93.2          & 75.4          & 92.5         \\ \hline
Non-local network\cite{non-local} & 77.7          & 93.3          & N/A           & N/A          \\ \hline
\end{tabular}
}
\vspace{-0.8em}
\label{tbl:bestresult}
\end{table}

\paragraph {Results.} The results are shown in Table~\ref{tbl:bestresult} (mean of 2 runs). 
On Kinetics, our proposed UniDual-Aux model yields a gain of 1.5\% over the already-strong performance achieved by R(2+1)D-152. Our model performs on-par with Non-Local Networks~\cite{non-local} which leverage a costly mechanism of nonlinear feature aggregation over the entire spatiotemporal volume. On the image task, the proposed model also outperforms the R2D-152 baseline model. We note that here we do not compare with state-of-the-art models on ImageNet since 1) our model uses a deflated R(2+1)D model, which has a smaller number of channels than the traditional ResNet model, and 2) we do not make use of data augmentation traditionally used to boost performance for image tasks. Our objective here is merely to show that our approach yields accuracy gains even when applied to high-capacity networks.

\paragraph{Visualization of UniDual.}

\begin{figure}[!htbp]
\vspace{-3mm}
\centering
     \begin{subfigure}[b]{0.95\textwidth}
         \centering
         \includegraphics[width=\textwidth]{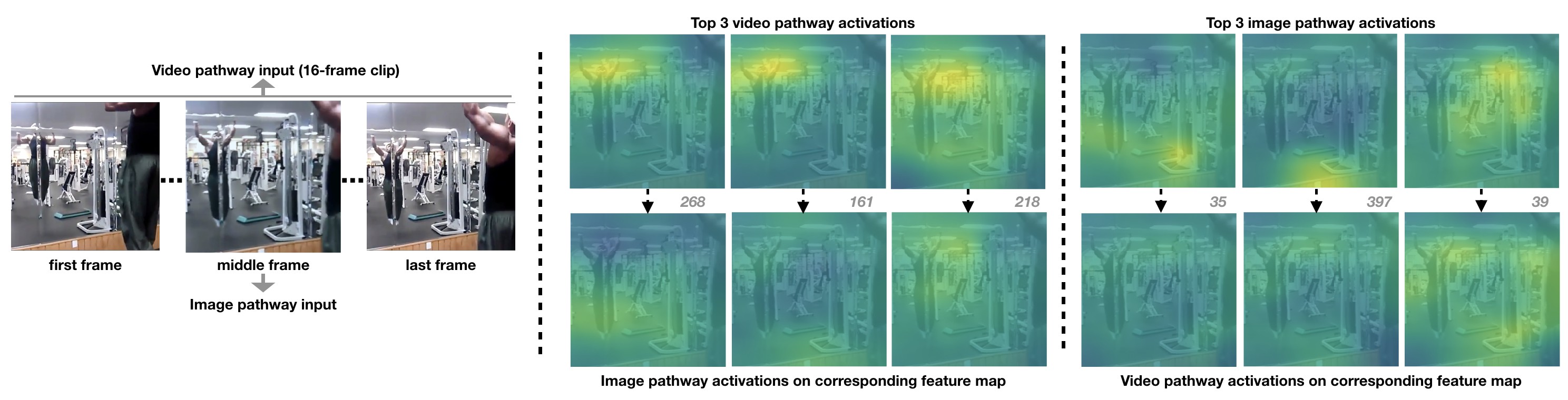}
         \label{fig:visualization}
     \end{subfigure}   
\vspace{-3mm}
     \caption{For a given video clip of a pull-up action, the video pathway (applied to the clip) and the image pathway (applied to the middle frame) capture different information in the dual point-wise layers. For each pathway we show the 3 most activated feature maps from the last residual block and the corresponding feature maps for the other pathway. The video pathway captures motion information (pull-up), while the image pathway focuses on appearance and object information (gym equipment). Grey numbers beside the feature maps indicate the channel index. } 
   \label{fig:visualization}
\vspace{-2mm}
\end{figure}

In Figure~\ref{fig:visualization}, we show examples of feature maps computed from the dual point-wise layers in the last residual block, using UniDual-152-Aux. For the given input clip (representing a pull-up action), we visualize the 3 most activated feature maps by the video pathway (top, middle). We also show the activations computed on the same feature channels by the image pathway using just the middle frame as input  (bottom, middle). It can be seen that the most activated feature maps from the video pathway focus on motion of the upper body, while the activations on the corresponding feature maps from the image pathway are spatially diffused and less peaked. However, if we look at the 3 most activated feature maps by the image pathway (top, right) we see strong activations on the gym equipment. This example suggests that the dual point-wise filters capture different information, with the temporal filters activating on motion, while the $1\times 1$ filters focus on objects or appearance cues.

\vspace{-0.6em}
\section{Conclusions}
\vspace{-0.6em}
\label{sec:conclusion}
Our work introduces a novel architecture aiming at capturing shared appearance cues in images and videos, while enabling the separate modeling of image-specific properties and temporal information in video. 
Our UniDual model accepts both images and video as input and activates different execution pathways depending on the input modality. We demonstrate that the strong performance of our model hinges on maintaining similar data statistics across all layers during optimization. We achieve this goal by means of auxiliary losses, one added to the image pathway and one added to the video pathway. Across all our experiments, our model achieves accuracy gains over 1) disjoint training, 2) pretraining followed by finetuning, and 3) traditional multi-task learning. In practical settings, our model enables the deployment of a single model to address different tasks over different modalities with the additional benefit of the improvements in accuracy. Future work will be focused on enhancing the scalability of the approach in order to enable the training of a unified model over a large number of different datasets and video/image tasks.


\bibliographystyle{ieee}
\bibliography{ieee_ref}

\clearpage


\section{More Visualization of {\em UniDual} Point-wise Filters}

\begin{figure}[!htbp]
\centering
     \begin{subfigure}[b]{1\textwidth}
         \centering
         \includegraphics[width=\textwidth]{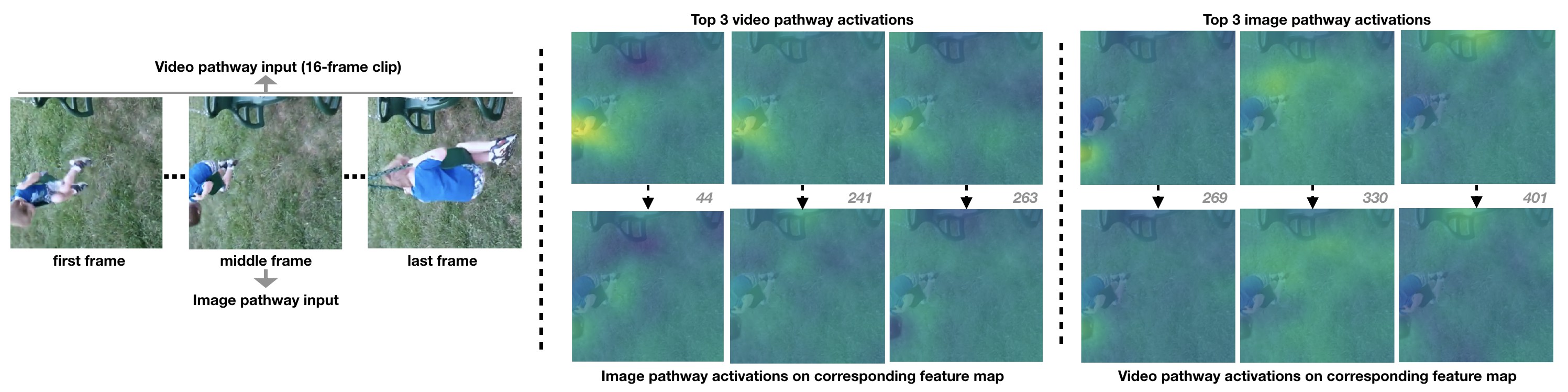}
         \caption{Swinging on something}
         \label{fig:swing}
     \end{subfigure}
     \hfill
     \begin{subfigure}[b]{\textwidth}
         \centering
         \includegraphics[width=\textwidth]{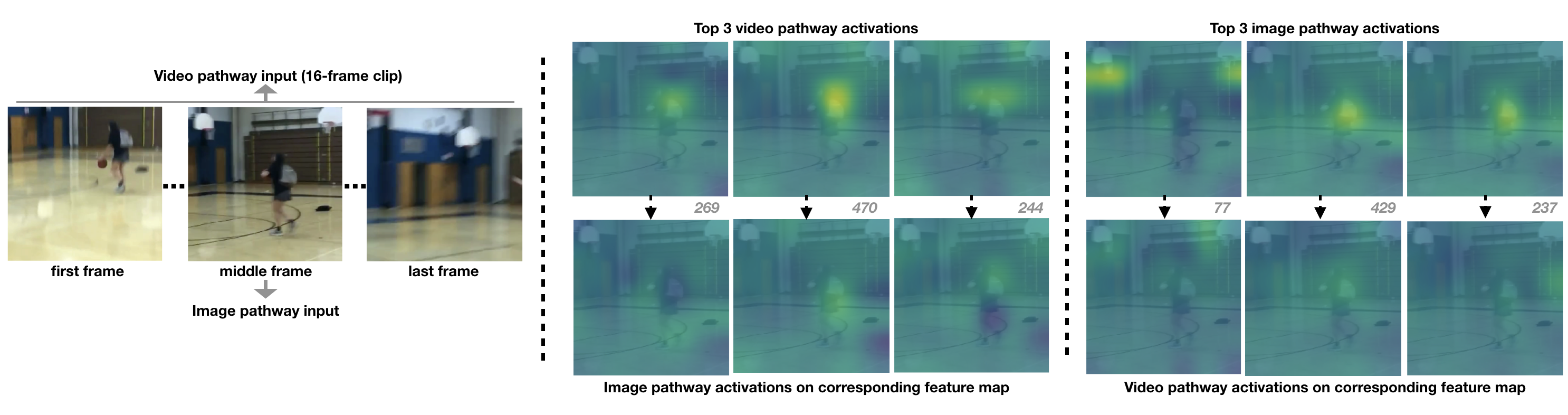}
         \caption{Shooting basketball}
         \label{fig:shoot}
     \end{subfigure}
     \hfill
     \begin{subfigure}[b]{1\textwidth}
         \centering
         \includegraphics[width=\textwidth]{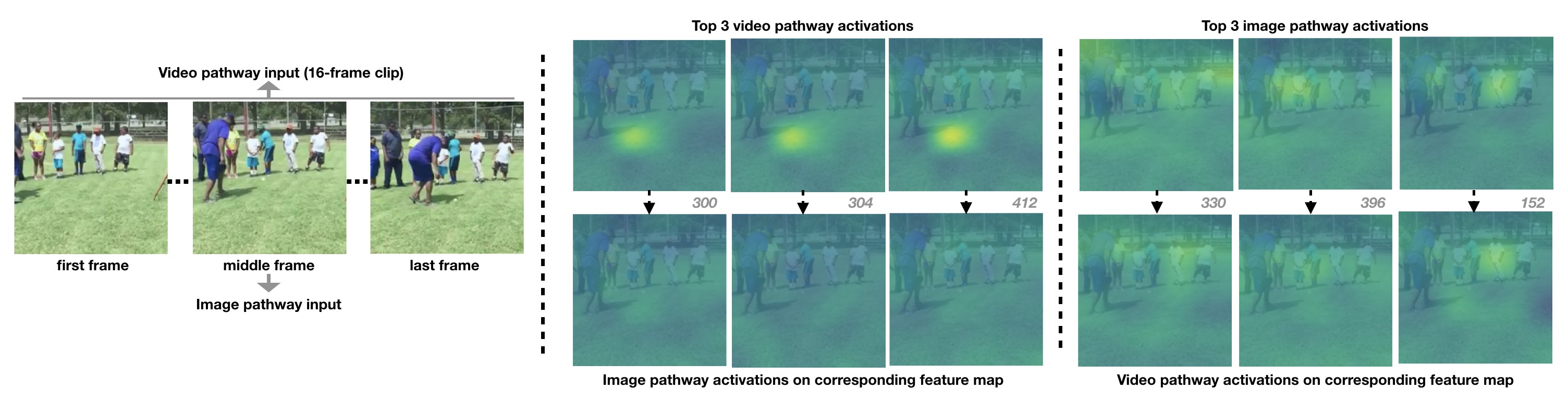}
         \caption{Catching or throwing baseball}
         \label{fig:throw}
     \end{subfigure}        
     \hfill
     \begin{subfigure}[b]{1\textwidth}
         \centering
         \includegraphics[width=\textwidth]{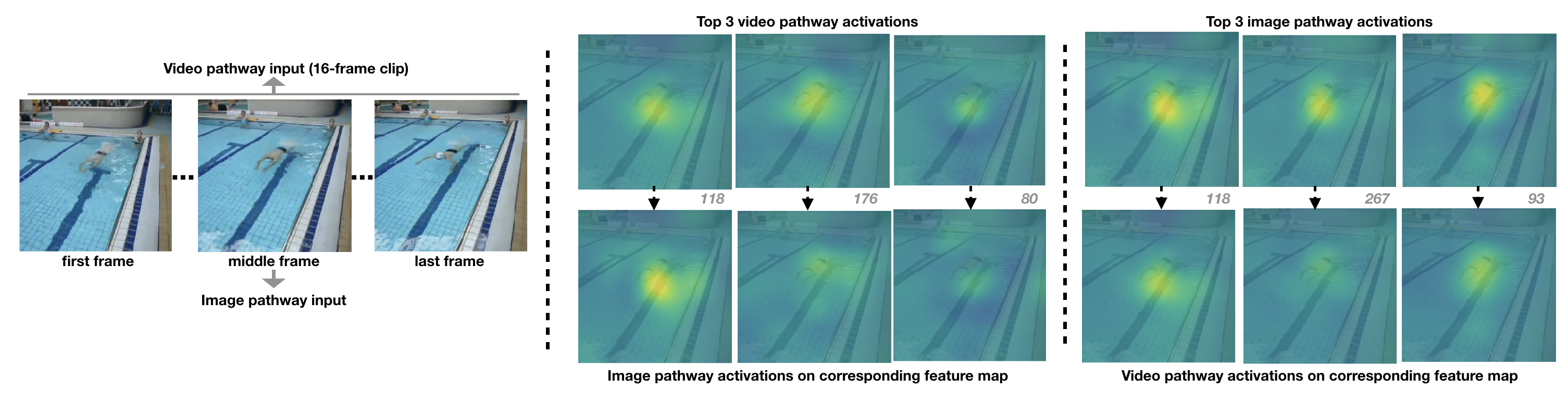}
         \caption{Swimming butterfly stroke}
         \label{fig:swim}
     \end{subfigure}    
   \caption{For a given input clip, the video pathway (applied to a 16-frame clip) and the image pathway (applied to the middle frame) capture different information in the dual point-wise layers.} 
   \label{fig:viz}
\end{figure}

In Figure~\ref{fig:viz}, we provide examples of feature maps computed from the dual point-wise layers in the third-to-last residual block, using UniDual-152-Aux. We choose the third-to-last residual block to show that in the middle of the network, the distinction between two pathways already emerges. Note that the receptive field is smaller in the feature maps we show here than that we show in the main paper (where we show the feature maps computed from point-wise layers in the last residual block).
 
We show 4 examples from 4 different actions on Kinetics in fig~\ref{fig:viz} (a)-(d). In each example, from left to right: we first show 3 images from the input clip (first, middle, and last), the top-3 most activated feature maps by the video pathway, and then the top-3 most activated feature maps by the image pathway. In video and image feature activation maps, we also visualize the corresponding feature maps for its dual path (bottom row).

As shown, the video pathway and the image pathway capture different information in the dual point-wise layers. For example, in Figure~\ref{fig:swing} with action of swinging, the video pathway focuses on the motion of the boy on the swing, while the activations on the corresponding feature maps from the image pathway are spatially diffused and less peaked. On the other hand, the image pathway focuses on the objects: the boy, the grass, and the other swing. In Figure~\ref{fig:shoot} with action of shooting basketball, video pathway focuses on the upper body of the moving person with the basketball, while image pathway focuses on both the basket and the person with the basketball. In Figure~\ref{fig:throw} with action of throwing a ball, the video pathway focuses on the action of ball thrown to the ground, while the image pathway focuses on the persons in the background.

However, Figure~\ref{fig:swim} with the action of swimming, we can see that the most activated channels for video pathway and image pathway are both Channel \#118. This shows that when the appearance information is strong, the video pathway still captures the same appearance information with the image pathway.

\clearpage

\end{document}